\documentclass[nohyperref]{article}
\usepackage[utf8]{inputenc}

\usepackage[accepted]{icml2022}

\title{CNT (Conditioning on Noisy Target): A new Algorithm for Leveraging Top-Down Feedback}
\author{}
\date{December 2021}

\usepackage{natbib}

\usepackage{wrapfig}
\usepackage{amsmath,amssymb,amsfonts,amsthm}

\usepackage[utf8]{inputenc} 
\usepackage[T1]{fontenc}    

\usepackage[colorlinks=true, citecolor=black]{hyperref}
\usepackage{url}            
\usepackage{booktabs}       
\usepackage{amsfonts}       
\usepackage{nicefrac}       
\usepackage{microtype}      
\usepackage{comment}

\usepackage{hyperref}       
\usepackage{url}            
\usepackage{booktabs}       
\usepackage{nicefrac}       
\usepackage{amssymb} 
\usepackage{amsmath}
\usepackage{amsthm}
\usepackage{graphicx}
\usepackage{subcaption}
\usepackage{caption}
\usepackage{color}
\usepackage{mathtools}
\usepackage[symbol]{footmisc}
\usepackage{tablefootnote}

\newcommand{\Zero}{\boldsymbol{0}}
\newcommand{\I}{\boldsymbol{\mathrm{I}}}
\newcommand{\x}{\boldsymbol{\mathrm{x}}}

\newcommand{\y}{\boldsymbol{\mathrm{y}}}

\newcommand{\w}{\boldsymbol{\mathrm{w}}}

\newcommand{\dr}{\mathrm{d}}

\icmltitlerunning{Conditioning on noisy target}

\begin{document}

\twocolumn[
\icmltitle{CNT (Conditioning on Noisy Targets): A new Algorithm for Leveraging Top-Down Feedback}



\icmlsetsymbol{equal}{*}

\begin{icmlauthorlist}
\icmlauthor{Alexia Jolicoeur-Martineau}{equal,x}
\icmlauthor{Alex Lamb}{equal,y}
\icmlauthor{Vikas Verma}{z1,z2}
\icmlauthor{Aniket Didolkar}{x}
\end{icmlauthorlist}

\icmlaffiliation{x}{Department of Computer Science, University of Montreal}
\icmlaffiliation{y}{Microsoft Research Redmond, USA}
\icmlaffiliation{z1}{Google Research, Brain Team}
\icmlaffiliation{z2}{Aalto University, Finland}

\icmlcorrespondingauthor{Alexia Jolicoeur-Martineau}{alexia.jolicoeur-martineau@mail.mcgill.ca}
\icmlcorrespondingauthor{Alex Lamb}{lambalex@microsoft.com}

\icmlkeywords{regularization, progressive learning, classification, imitation learning}

\vskip 0.3in
]



\printAffiliationsAndNotice{\icmlEqualContribution} 

\begin{abstract}
We propose a novel regularizer for supervised learning called Conditioning on Noisy Targets (CNT). This approach consists in conditioning the model on a noisy version of the target(s) (e.g., actions in imitation learning or labels in classification) at a random noise level (from small to large noise). At inference time, since we do not know the target, we run the network with only noise in place of the noisy target.  CNT provides hints through the noisy label (with less noise, we can more easily infer the true target). This give two main benefits: 1) the top-down feedback allows the model to focus on simpler and more digestible sub-problems and 2) rather than learning to solve the task from scratch, the model will first learn to master easy examples (with less noise), while slowly progressing toward harder examples (with more noise).
\end{abstract}

\section{Introduction}

\begin{figure}
    \centering
    \includegraphics[width=1\linewidth]{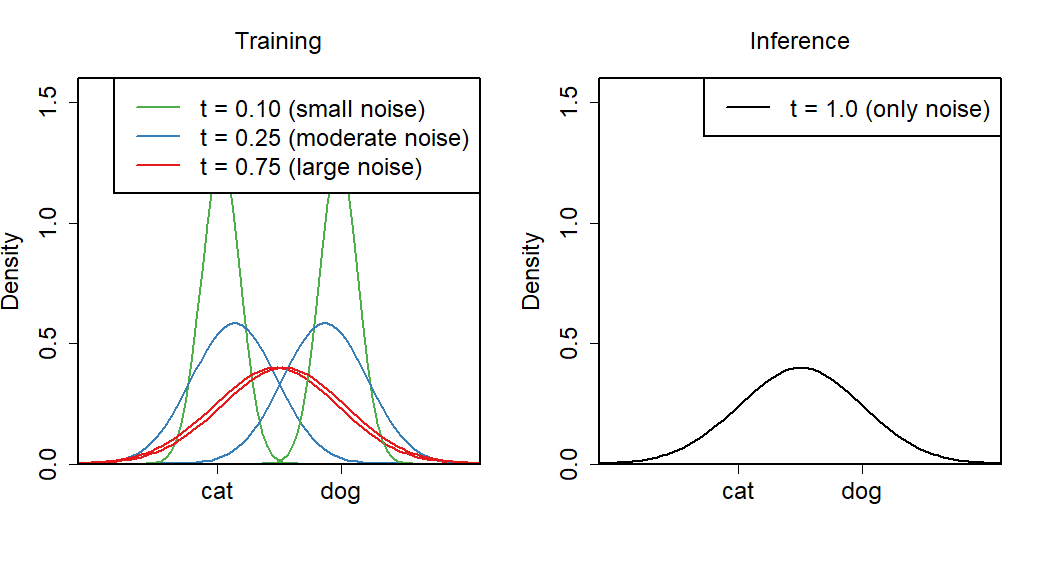}
    \caption{This figure illustrates CNT in a ``cat'' versus ``dog'' classification problem. During training, we help the model by giving it a hint which corresponds to the noisy target. When the noise is small, the model can "cheat" and tell apart the true target. But as the noise increases, the model is given less hint and must rely more and more on the input image. During inference, the target is unknown, so the model is only given noise (no hint).
    } 
  \label{fig:flagship}
  \vspace{-4mm}
\end{figure}



Astonishingly, deep learning can solve complex high-dimensional problems from a randomly initialized neural network without any prior experience/knowledge. However, the path from a random network to one that can effectively solve a problem is long and arduous. Rather than learning to solve a complex problem from scratch, humans tend to learn progressively: by starting from easy tasks and slowly increasing their difficulty.

Progressive learning methods have also been used in deep learning due to their benefits in improving model generalization \citep{fayek2018transferability}. Curriculum learning consists in ordering the samples of the dataset in order of difficulty to help learn progressively \citep{bengio2009curriculum}. Gradual transfer learning consists in moving from easier to harder datasets \citep{fayek2018transferability}. Progressive growing consists in growing the neural network over time to replicate neural growth \citep{terekhov2015knowledge, rusu2016progressive} or to solve easier tasks before solving harder tasks \citep{karras2017progressive}.

Although progressive learning methods can improve generalization, these methods are often non-trivial to incorporate. Progressive growing requires a complex architecture design and extensive hyperparameter tuning. Curriculum learning requires access to a model that estimates the difficulty of each sample so that we can train a network on samples of increasing difficulty. Gradual transfer learning requires multiple datasets sorted by difficulty and extensive architecture/hyper-parameter tuning. 


In this work, we seek to help the model progress from easier to harder tasks by providing it {\em hints} during training. Ideally, we would like such a method to be easy to incorporate within any given problem and not require hyper-parameter tuning. Given the strong evidence that progressive learning methods can improve generalization \citep{bengio2009curriculum, fayek2018continual}, we believe that providing hints to the network would improve generalization in supervised learning problems and be easier to incorporate than most existing progressive learning methods.

This raises the question: {\em How can one provide hints about the solution to a neural network?} We propose to give hints to the neural network by conditioning the neural network on a noisy version of the target(s) (e.g., actions in imitation learning or labels in classification). When adding very little noise to the target, the true target can be easily deduced by the neural network thus making the problem easier. Meanwhile, when adding a large amount of noise to the target, the true target is harder to infer from the noisy target. This provides a path for the neural network to learn to solve easier problems (noisy target with less noise) before harder problems (noisy target with more noise). 


At the same time, we may consider why a noisy-version of the target is appropriate as a ``hint'' for training.  An intuitive reason is that the target contains high-level and semantically meaningful top-down information, which is relevant for perception.  The role of top-down feedback in perception has been studied in the cognitive science literature \citep{Rauss2013,Kinchla1979} as a way of improving robustness to challenging or unreliable bottom-up signals.  For example, a person may be able to navigate in a dark room that they are familiar with by leveraging priors about where objects are expected to be present.  In our case, a model may be able to more easily make predictions early in training when it has access to noisy-targets, which indicate a handful of target values which are the most likely.  

Adding noise to the target and conditioning the model on this noisy target should be straightforward. However, progressively increasing the noise according to a schedule over time is complicated; doing so would require heavy hyper-parameter tuning and move us away from our goal of having a simple tuning-free method.

Instead of a progressive approach, we take a {\em multi-scale} approach. An example of a multi-scale approach would be the Multi-Scale Gradient (MSG) approach by \citet{karnewar2019msg}. Instead of progressively growing an architecture to generate images from low-resolution to high-resolution \citep{karras2017progressive}, MSG proposed to generate images from all resolutions simultaneously. MSG is now used in StyleGAN2 \citep{karras2020analyzing} instead of progressive growing because it enhances the stability and quality of the generated samples. The main benefit of a multi-scale approach is that the model is given gradients at all levels of difficulty, which means that it cannot unlearn solving the simpler tasks (generating lower-resolution images with MSG) after learning to solve the harder tasks (generating higher-resolution images with MSG).

Given the benefit and simplicity of the multi-scale approach, we use it instead of progressively increasing the noise. First, we sample uniformly from a continuous noise level ($t \sim \mathcal{U}([0,1])$) where $t=0$ means no noise, and $t=1$ means only noise. Then, we inject an embedding on the noise level and the noisy label (hint) into the network. At test time, instead of sampling from the uniform distribution, we take $t=1$, which means we have no information about the label and must start without any hint. 

\begin{figure}
    \centering
    \includegraphics[width=1\linewidth]{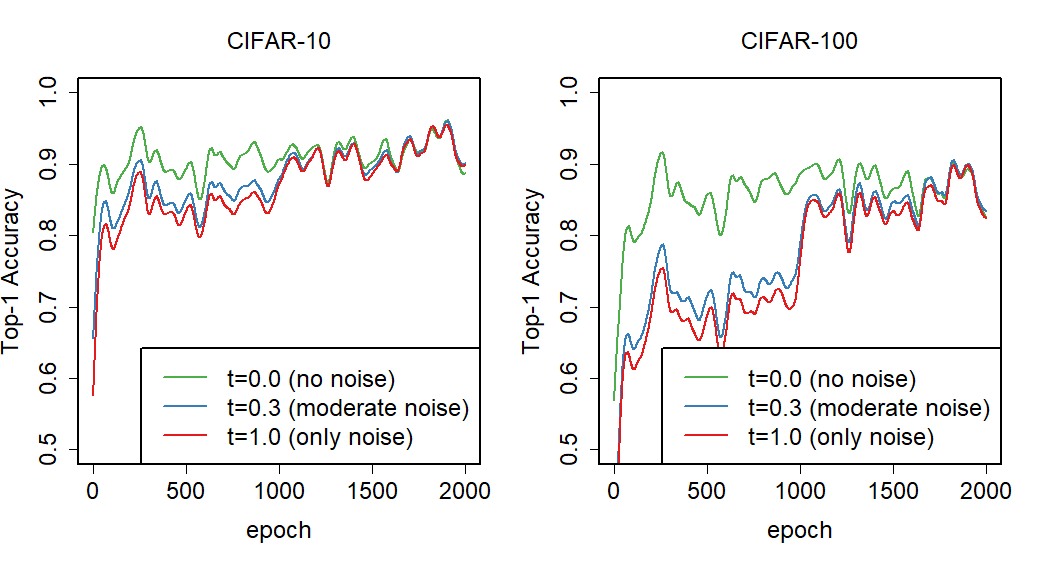}
    \caption{This figure highlights the multi-scale progressive learning of CNT. It shows the Top1 accuracy of the last mini-batch at the end of each training epoch. The model learns to master easier examples (with less noise) before mastering harder examples (with large noise).} 
  \label{fig:noiselevel} 
\end{figure}

We call this technique Conditioning on Noisy Targets (CNT). CNT helps the network learn from multiple difficulties simultaneously and provides a continuous path for the model to progress from maximum hints (no noise) to no hints (only noise). We illustrate our approach in Figure \ref{fig:flagship} and highlights how it learn progressively in a multi-scale fashion in Figure \ref{fig:noiselevel}.  Our contribution is the Conditioning on Noisy Target (CNT) approach, a multi-scale task-agnostic regularizer providing top-down feedback to the neural network to improve the efficiency of learning. This approach is easy to implement as it only requires adding a small embedding network (for the noisy output and noise-level) and injecting the embedding in conditional normalization layers, which replace the usual normalization layers; no other changes are needed. Our approach can be applied to any neural network on any supervised learning problem and requires no hyper-parameter tuning. We test the validity of the proposed method on various different tasks like using image labels as a target for image classification problems, using the information about actions as a target for reinforcement learning problems.

\section{From Supervised learning to CNT}

We want to solve a supervised learning problem: 
\begin{equation}
    \min_{\theta} \mathbb{E}_{p(\x,\y)} \left[ L(f_{\theta}(\x), \y) \right],
\end{equation}
where $L$ is an arbitrary loss function, $f_{\theta}$ is a neural network with parameters $\theta$, $(\x,\y)$ is a pair of data and target sampled from a dataset.

For generality, we consider $\y \in \mathbb{R}^{CL}$ to be a real-valued vector where $C$ is the number of classes and $L$ is the number of labels.  In the case of multi-class classification, we use the one-hot vector representation as $\y$.  In the multi-label case, we concatenate the one-hot vectors for all of the labels together to construct $\y$.  

With most loss functions used in supervised learning (e.g., least-squares and cross-entropy), the optimal model returns the conditional expectation of $\y$ given $\x$:
\begin{equation}
    f_{\theta}^{*}(\x) = \mathbb{E}_{p(\y|\x)} \left[ \y | \x \right].
\end{equation}

\subsection{Noisy target}

Instead of having $\y$ be the target, we now want to replace it with a noisy target $\{ \y(t) \}_{t=0}^1$, so that $\y(0)$ is the true target and $\y(1)$ is pure noise. Let $(\x, \y(0))$ be a pair of data/target sampled from the data distribution. Any type of time-dependent noise could be used. Following the recent success of score-based diffusion models \citep{ho2020denoising,song2020scorebased}, we chose to gradually corrupt the target over time using a diffusion process: 
\begin{equation}\label{eq:forward}
\dr\y = f(\y,t)\dr t + g(t)\dr \w,
\end{equation}
where $f(\y,t): \mathbb{R}^d \times \mathbb{R} \to \mathbb{R}^d$ is the drift, $g(t): \mathbb{R} \to \mathbb{R}$ is the diffusion coefficient and $\w(t)$ is the Wiener process indexed by $t \in [0,1]$. 

The functions $f$ and $g$ are chosen so that $\y(0)$ is the true label ($f(\y,0))=g(\y,0)=0$) and $\y(1)$ is independent from $\y(0)$ so that $\y(1)$ can be sampled from even when $\y(0)$ is unknown (at inference time). Given the good empirical results from the Variance Preserving (VP) process in score-based diffusion models \citep{song2020scorebased}, we use it as our diffusion process; it is defined as follows:
\begin{equation}
    \dr \y = - \frac{1}{2} \beta(t) \y \dr t + \sqrt{\beta(t)} \dr\w.
\end{equation}

We can create noisy targets from the VP process by sampling from the following distribution:
\begin{equation}\label{eqn:vp_marginal}
    \y(t)|\y(0) \sim \mathcal{N}(\y(0)~e^{-\frac{1}{2} \int_{0}^t \beta(s) \dr s}, (1- e^{-\int_{0}^t \beta(s) \dr s})~ \I ),
\end{equation}
where $\beta(t) = \beta_{min} + t \left( \beta_{max} - \beta_{min} \right)$,  $\beta_{min} = 0.2$ and $\beta_{max}=20$.

Thus, as desired, $\y(0)$ is approximately the target and $\y(1)$ is approximately distributed as $\mathcal{N}(\Zero, \I)$ and does not depend on $\y(0)$.

Instead of using a diffusion process, it is also possible to use a different type of noise. For example, one could use Laplace distributed noise instead of Gaussian distributed noise. The Laplace distribution has heavier tails while having a higher concentration at its mean. The high concentration at the mean may make it easier to know how close the hint is to the truth. To add Laplace noise, we follow equation \ref{eqn:vp_marginal} but replace the Gaussian distribution by the Laplace distribution: \begin{equation}\label{eqn:vp_laplace_marginal}
\scalebox{0.90}{
    $\y(t)|\y(0) \sim Laplace \left( \y(0)~e^{-\frac{1}{2} \int_{0}^t \beta(s) \dr s}, \sqrt{1- e^{-\int_{0}^t \beta(s) \dr s}} \right)$.
    }
\end{equation}
Our paper focuses on the Gaussian VP Process, but we also study its Laplace equivalent in a few settings.  

\subsection{CNT (Conditioning on Noisy Target)}

Rather than solving the supervised learning problem using only $\x$, we now want to further condition on the noisy target:
\begin{equation}\label{eqn:cont}
    \min_{\theta} \mathbb{E}_{p(\y(t)|\y(0)),p(\x,\y(0))} \left[ L(f_{\theta}(\x | \y(t), t), \y) \right],
\end{equation}
where $f_{\theta}$ is now a neural network taking $\x$, $\y(t)$, and $t$ as inputs.

At the optimum, the neural network is now the conditional expectation of $\y$ given $\x$, $\y(t)$, and $t$:
\begin{equation}
    f_{\theta}^{*}(\x | \y(t), t) = \mathbb{E}_{p(\y)} \left[ \y | \x, \y(t), t \right].
\end{equation}

At $t=0$, the problem is already solved and the the optimal network returns back the target:
\begin{equation}
    f_{\theta}^{*}(\x | \y(0), t=0) = \y(0).
\end{equation}

At $t=1$,the optimal network ignores the noisy label since it independent of the label:
\begin{equation}\label{eqn:ind}
    f_{\theta}^{*}(\x | \y(1), t=1) = \mathbb{E}_{p(\y)} \left[ \y | \x \right].
\end{equation}
At inference, since we do not know the true label, we use $t=1$ to recover our original expected value of $\y$ given $\x$. 

Since we sample from a random noisy $\y(1)$, it might appear sensible to average over multiple random draws of $\y(1)$ at inference; however, as shown above, the noise is completely independent of the true label at $t=1$, and the model should ignore the noise by converging to the expected value of $\y$ given $\x$. We found no benefit from using multiple random draws; it only made inference slower. We illustrate CNT with the VP process in Figure \ref{fig:flagship}. During training, we condition on the $\y(t)$ with a random noise-level $t$ as shown in Equation \ref{eqn:cont}. At inference time, we do not know the true label, so we only provide pure noise $\y(1) \sim \mathcal{N}(0,I)$ and the output should return the expected value of $\y$ given $\x$.

\begin{figure}
    \centering
    \includegraphics[width=1\linewidth]{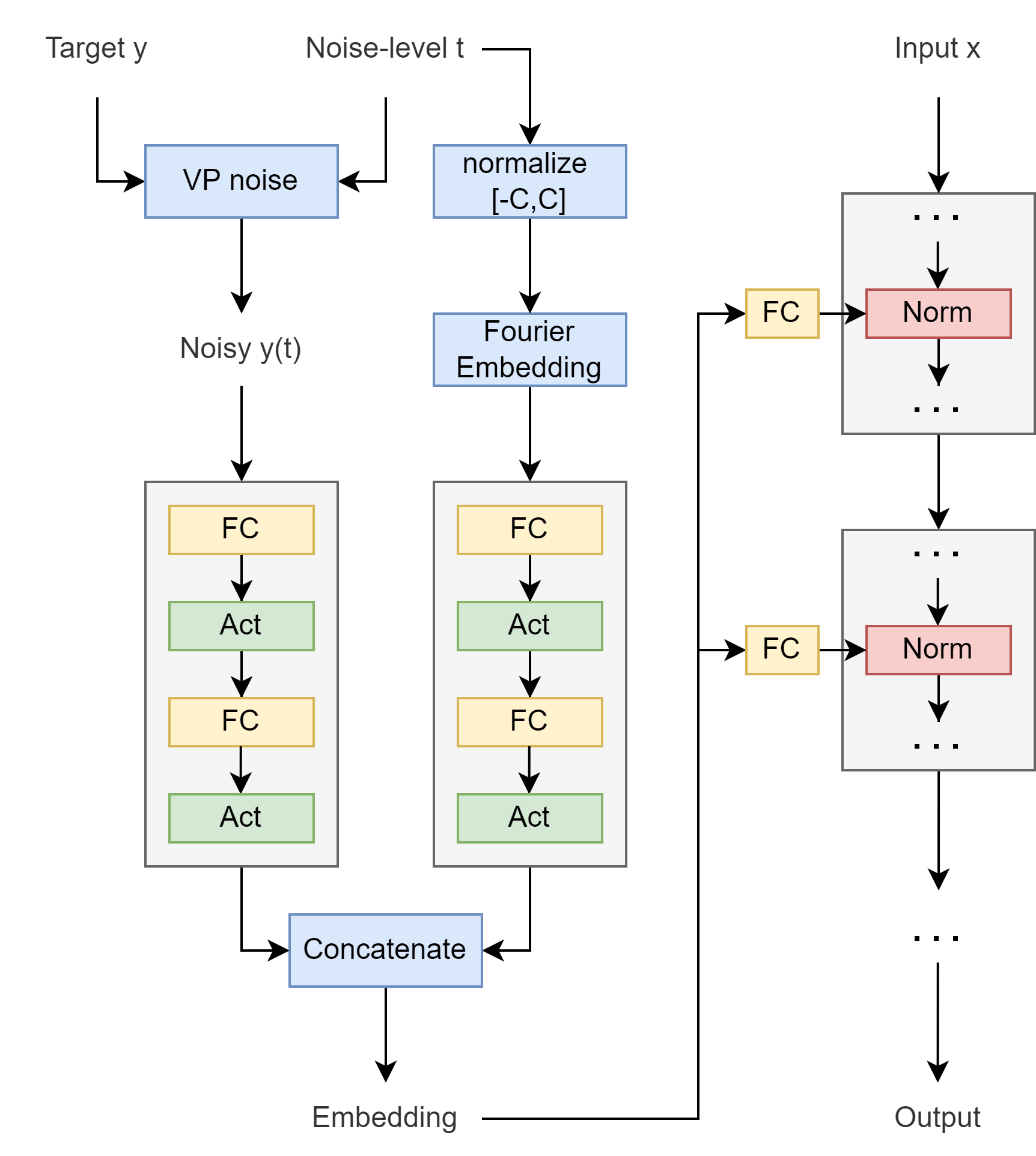}
    \caption{This figures demonstrate how to condition on the noisy target and noise-level in CNT.
    } 
  \label{fig:arch}
  \vspace{-8mm}
\end{figure}

\subsection{Implementation}

There are many possible ways of conditioning on the noisy target $\y(t)$. A very popular way of input conditioning is through conditional normalization layers \citep{huang2017arbitrary, de2017modulating, karras2019style,dhariwal2021diffusion}; this is the approach we use.

Our approached for conditioning on the noisy target is explained in Figure \ref{fig:arch} and below. We start by processing $t$ through a Fourier positional embedding \citep{tancik2020fourier} in order to increase the high-frequency signal. Then, we separately process both $\y(t)$ and the Fourier processed $t$ through small fully-connected networks. Then, we concatenate both outputs to produce an overall embedding for $(t, \y(t))$. After each normalization layer, instead of using learned fixed weights and bias, we learn noise-conditional weights and bias, which are linear projections of the embedding. 

The model conditional on the noisy-target has more parameters than a unconditional model due to the embedding parameters and the weights of the linear projections. However, given a fixed $(t, \y(t))$, the network is equivalent to an unconditional neural network. Furthermore, the model should learn to ignore $\y(t)$ when $t=1$ given that the noisy target contains zero information about the target at that level of noise. This relatively simple process can be adapted to any network, even those without normalization layers, by simply applying the conditional weights and bias at different locations through the network.

\section{Related Work}

Several lines of research explore topics related to the CNT technique.  

\paragraph{Leveraging Top-Down Feedback.} One can learn representations in a bottom-up or top-down manner. Bottom-up refers to the scenario when the low-level (sensory) representations  modulate high level (conceptual) representations. Top-Down refers to the scenario when during information processing, information from high-level representations modulate lower-level representations. Exploring the combination of top-down and bottom-up signals in network architectures has an important
history in deep learning \citep{theeuwes2010top, perez2018film,  anderson2018bottom, carreira2016human, mittal2020learning, lamb2021neural}. The goal has been to explore the value of combining top-down mechanisms for efficient learning, robustness to noise or distractors and hence achieving efficient transfer across related tasks by exploiting the task structure. The proposed regularizer is also relevant to models which incorporate feature wise modulation based on some conditioning information. The conditioning information is used to influence the computation done by a neural network.  For example. conditioning information can take the form of ``language instruction'' to modulate the processing of a neural network learning a representation of an visual image \cite{perez2018film}. It can also take the form of goal information that is used to condition the policy as in goal based policies in reinforcement learning \cite{schaul2015universal}. In the proposed framework, we use the noisy version of the ``target'' as the conditioning information, and we anneal the noise level to force the network to iteratively learn representations by focussing on simpler patterns in the data. 

\paragraph{Curriculum and Progressive Learning} Training deep networks on examples which progress on a curriculum from easier to more difficult examples has been a successful strategy, with benefits for both ease of optimization and generalization \citep{bengio2009curriculum}.  \cite{soviany2021curriculum} identifies dozens of successful applications of curriculum learning across several domains, including computer vision, speech recognition, medical image analysis, and natural language processing.  At the same time, successful use of a curriculum requires two challenging hyperparameters: a schedule for when to switch from easier to more difficult examples and an ordering of examples into easier and harder examples \citep{hacohen2019power}.  In some sequence problems, it is relatively easy to design such a curriculum.  For example, sorting a short list of numbers is easier than sorting a longer list of numbers.  In our technique, we construct a continuum of easier and harder set of examples which we train jointly, with lower noise-levels providing easier examples and higher noise-levels leading to harder examples.  

\paragraph{Injecting Noise into Deep Models} Training deep neural networks with noise has proved to be a surprisingly successful regularizer.  The dropout technique randomly sets a percentage of the units in a deep network to zero on each forward pass \citep{srivastava2014dropout} and has been highly successful across a variety of domains.  Adding noise to a neural network's weights has also been shown to be useful in reinforcement learning \citep{fortunato2017noisy}.  Injecting noise into the hidden states of a neural network using a variational bottleneck has also been shown to be a successful regularizer \citep{alemi2016vib}.  CNT shares the idea of injecting noise into the hidden states of a deep network (through conditional normalization layers), and we found that this by itself often improved results over the baseline.  However, an important difference with CNT is that conditioning is performed on a noisy target with a variable amount of noise, and the model is also conditioned on the noise level being injected.  Because the target contains information which is useful for solving the task, the network has a natural inductive bias to leverage the noisy target.  

\paragraph{Denoising Generative Models} CNT can be seen as a model for denoising the target $\y$ conditioned on the input $\x$.  Denoising (typically done over multiple steps) has frequently been used to construct generative models $p(\x)$ or conditional generative models $p(\x | \y)$.  \cite{bengio2013generalized} showed that denoising autoencoders can be used as generative models by running multiple steps of denoising during the sampling process.  Work on deep diffusion, which involves learning to denoise from multiple levels of noise during both training and sampling has led to competitive conditional and unconditional generative models \citep{ho2020denoising,ho2021cascaded}.  Our work differs from these other works in that we focus exclusively on the discriminative task of modeling $p(\y | \x)$.  Moreover this distribution is usually nearly uni-modal (or has only a few modes), which means that the denoising process can be completed perfectly in only a single step.  This removes the need to run multiple steps during sampling and also removes the need to draw multiple samples.  

\section{Experiments and Results}

To show the generalization benefits of CNT in supervised learning, we test CNT over a wide range of problems: image classification, relational reasoning (Sort-of-CLEVR), equilateral vs. non-equilateral shape classification, and reinforcement learning.  

To ensure that the benefit we see in CNT comes from the label information in the noisy target and not simply from noise conditioning, we always test for noise-only conditioning: $\beta_{min} = \beta_{max}$ in the VP process so that, for any $t$, it is pure standard Gaussian noise. It is well known that regularization through injecting noise can sometimes be beneficial; thus, it is important to compare CNT to only-noise.

Preliminary analyses showed that having the Mish activation function in the CNT embedding generally leads to slightly better results (accuracy in classification and Shapes) than using ReLU. Thus, when comparing only-noise and CNT to baseline models using ReLU, we use still use Mish inside the embedding and in the linear projections for the conditional normalization. Having a continuous embedding ensures that the embedding does not change significantly with respect to small changes in the noise-level $t$, and we believe that this is beneficial.

\subsection{Classification}

We test CNT for classification with the following datasets: CIFAR-10 \citep{krizhevsky2009learning}, CIFAR100 \citep{krizhevsky2009learning}, TinyImageNet \citep{chrabaszcz2017downsampled}, and ImageNet \citep{deng2009imagenet}. We use a basic ResNet architecture \citep{he2016deep} for all out models. 

We optimize the cross-entropy loss function. As regularizations, we use dropout \citep{srivastava2014dropout} with $p=0.1$ \citep{srivastava2014dropout} and the mixup data augmentation \citep{zhang2017mixup} with $\alpha=1$. The models are trained with Stochastic Gradient Descent (SGD) with learning rate $0.1$, momentum $0.9$, and weight decay $0.0001$. The learning rate is scheduled to decrease by a factor of 10 at $50\%$ and $80\%$ of the training time. TinyImagenet models are trained for 1000 epochs, while other models are trained for 2000 epochs. To reduce variance, we train on all settings three times and report the mean and standard deviation. Results are shown in Table \ref{tab:class1}.

\begin{table}
    \small
	\caption{Classification: Test accuracy mean (standard deviation) from 3 seeds}
	\label{tab:class1}
	\centering
	\resizebox{.48\textwidth}{!}{%
	\begin{tabular}{cccc}
		\toprule
		Model & Baseline & only-noise & CNT  \\
		\cmidrule(){1-4}
		\multicolumn{4}{c}{TinyImageNet} \\
		\cmidrule(){1-4}
		ResNet18 - Mish & 63.77 (0.20) & 63.94 (1.02) & {\fontseries{b}\selectfont65.32} (0.59) \\
		ResNet18 - ReLU & {\fontseries{b}\selectfont 65.31} (0.05) & 64.38 (0.00) \makebox[0pt]{\footnotemark} & 64.87 (0.80) \\
		\cmidrule(){1-4}
		\multicolumn{4}{c}{CIFAR-100} \\
		\cmidrule(){1-4}
		ResNet18 - Mish & 79.21 (0.36) & 79.29 (0.84) & {\fontseries{b}\selectfont 80.23} (0.67) \\
		ResNet18 - ReLU & 79.93 (0.30) & 79.72 (0.58) & {\fontseries{b}\selectfont80.33} (0.90)  \\
		\cmidrule(){1-4}
		\multicolumn{4}{c}{CIFAR-10} \\
		\cmidrule(){1-4}
		ResNet18 - Mish & 96.74 (0.06) & {\fontseries{b}\selectfont 96.81} (0.15) & 96.70 (0.07) \\
		ResNet18 - ReLU & 96.63 (0.15) & {\fontseries{b}\selectfont 96.79} (0.07) & 96.70 (0.11) \\
		\bottomrule
	\end{tabular}
	}
\end{table}
\footnotetext{Some models collapsed}

\begin{table}
    \small
	\caption{Low-capacity classification: Test accuracy mean (standard deviation) from 3 seeds}
	\label{tab:class2}
	\centering
	\resizebox{.48\textwidth}{!}{%
	\begin{tabular}{cccc}
		\toprule
		Model & Baseline & only-noise & CNT  \\
		\cmidrule(){1-4}
		\multicolumn{4}{c}{CIFAR-100 ($ch=64$)} \\
		\cmidrule(){1-4}
	    ResNet9 & 72.11 (0.80) & 71.50 (4.33) & {\fontseries{b}\selectfont 74.23} (0.17) \\
		\cmidrule(){1-4}
		\multicolumn{4}{c}{CIFAR-100 ($ch=8$)} \\
		\cmidrule(){1-4}
	   ResNet9 & 44.45 (0.22) & 53.71 (0.89) & {\fontseries{b}\selectfont 54.19} (0.86) \\
	   ResNet18 & 57.94 (0.36) & 63.62 (0.81) & {\fontseries{b}\selectfont 63.76} (0.51) \\
		\cmidrule(){1-4}
		\multicolumn{4}{c}{TinyImagenet ($ch=64$)} \\
		\cmidrule(){1-4}
	    ResNet9 & 45.45 (0.10) & {\fontseries{b}\selectfont 53.50} (0.21) & 52.65 (0.41) \\
		\cmidrule(){1-4}
		\multicolumn{4}{c}{TinyImagenet ($ch=8$)} \\
		\cmidrule(){1-4}
	    ResNet9 & 19.05 (0.79) & 25.34 (0.53) & {\fontseries{b}\selectfont 33.45} (0.31) \\
	    ResNet18 & 34.46 (1.32) & 39.51 (0.60) & {\fontseries{b}\selectfont 41.64} (2.51) \\
		\bottomrule
	\end{tabular}
	}
\end{table}

From Table \ref{tab:class1}, we see that CIFAR-10 shows better results for only-noise, CIFAR-100 for CNT, and TinyImageNet for either CNT or baseline. Thus, only-noise and CNT almost always perform better than baseline, albeit by a small margin. The fact that only-noise is sometimes better (in CIFAR-10) suggests that adding noise may have its own benefits.

In the next set of experiments, we show that the benefits of CNT become larger and more consistent when the classification models have low capacity. To test the effect of CNT in a low-capacity model, we train the previous models with only 8 channels or/and with half the number of layers. These experiments use mixup $\alpha=0.5$. Results are shown in Table \ref{tab:class2}. From \ref{tab:class2}, we see that CNT almost always leads to higher accuracy than the other methods except for ResNet9 with TinyImageNet, which obtains slightly better accuracy with only-noise. This provides evidence that conditioning on the noisy target can be very beneficial, especially when the model struggles due to low capacity.


\subsection{Detecting Equilateral Shapes}

\begin{table}
    \small
	\caption{Equilateral Shape Detection: Test accuracy mean (standard deviation) from 3 seeds}
	\label{tab:shape}
	\centering
	\resizebox{.48\textwidth}{!}{%
	\begin{tabular}{cccc}
		\toprule
		Model & Baseline & only-noise & CNT  \\
		\cmidrule(){1-4}
		\multicolumn{4}{c}{Shapes} \\
		\cmidrule(){1-4}
	     CaffeNet - Mish & 83.57 (2.13) & 85.03 (0.80) & {\fontseries{b}\selectfont 85.77} (0.21) \\
	     CaffeNet - ReLU & 85.23 (0.40) & 86.03 (0.31) & {\fontseries{b}\selectfont 86.37} (0.65) \\
	     ResNet18 - Mish & 96.63 (0.15) & 96.77 (0.06) & {\fontseries{b}\selectfont 96.80} (0.10) \\
	     ResNet18 - ReLU & 96.57 (0.23) & 96.83 (0.21) & {\fontseries{b}\selectfont 96.87} (0.06) \\
		\bottomrule
	\end{tabular}
	}
\end{table}

We test CNT on the task of detecting geometrical shapes. We build on the equilateral triangle detection task from \cite{Ahmad2009EquilateralTA}. In addition to triangles, we also add squares and rectangles to our task. Each image in the dataset is of size $64 \times 64$. Each vertex of a geometrical shape is represented by a closely formed cluster of points centered at a randomly chosen coordinate in the image. For triangles, we add three such clusters to the image and for quadrangles, we add four such clusters to the image. We ensure that the positioning of these clusters satify the geometrical properties of the shape they represent. For example, we ensure that the clusters are equidistant for the equilateral triangle and the square. Also, we ensure that the sides for the quadrangles are perpendicular to each other.  



This task is framed as a classification task where we have two classification heads - One denotes whether the given shape has 4 sides or 3 sides and the second denotes whether the given shape has equal length sides or not. We use a basic ResNet architecture \citep{he2016deep} and CaffeNet \citep{jia2014caffe}. We use dropout \citep{srivastava2014dropout} with $p=0.1$ \citep{srivastava2014dropout} in the ResNet architecture. We incorporate noise by feeding the noisy label through the normalization layers in each architecture similar to Figure \ref{fig:arch}.  We minimize a binary cross entropy loss function for both classification heads. The models are trained for 200 epochs with Stochastic Gradient Descent (SGD) with learning rate $0.1$, momentum $0.9$, and weight decay $0.0001$. The learning rate is scheduled to decrease by a factor of 10 at $50\%$ and $80\%$ of the training time. To reduce the variance, we train each model three times and report the mean and standard deviation. For the result, we report the results for the second classification head i.e. whether the given shape has equal length sides or not.


\begin{figure}
    \vspace{-3mm}
    \centering
    \includegraphics[width=0.98\linewidth]{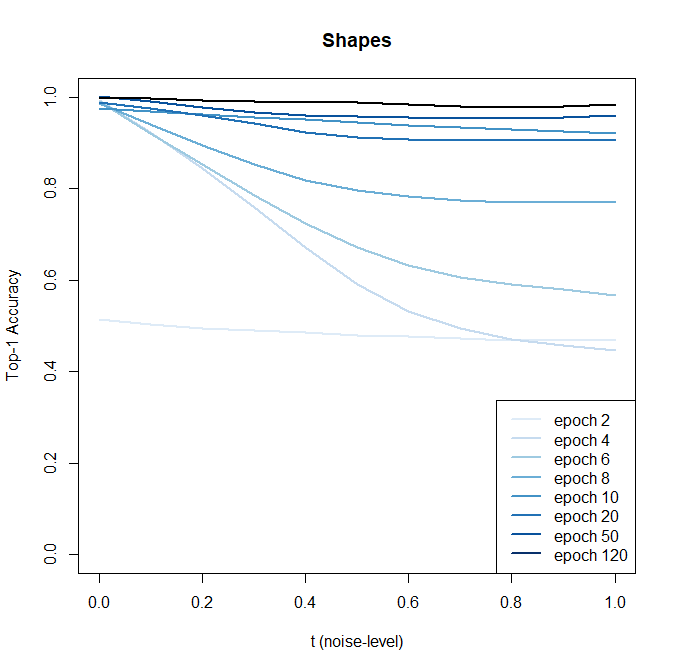}
    \caption{On the equilateral shape classification task, we found that training converges in very few epochs at low noise levels.  At higher noise levels, training requires more epochs to converge.  Moreover, we observe a smooth transition between these two regimes, with moderately fast convergence at intermediate noise levels.  }
    \label{fig:shapes_noise}
    \vspace{-5mm}
\end{figure}

We report the results in Table \ref{tab:shape}, we see that CNT consistently leads to higher accuracy in all settings. The effect of CNT is more significant with CaffeNet, probably because this architecture has a lower capacity than ResNet.

\subsection{Relational Reasoning (Sort-of-CLEVR)}

Generalization on relational reasoning tasks has often been difficult for deep supervised networks, which has been partially successfully addressed through the use of curriculum learning to encourage learning simpler skills which can be re-used and re-purposed to solve more complex tasks which require multiple skills.  For example, if a model needs to solve a relational question such as: ``What color is the shape to the left of the triangle?'', the model needs to be able to recognize colors, detect shapes, and understand relative positioning of shapes.  If a model masters all of these simpler individual skills it will generalize better when solving novel composite tasks which use many different skills.  

We evaluated CNT on the Sort-of-CLEVR relational reasoning benchmark \citep{NIPS2017_7082}. Each image in the sort-of-clevr benchmark is of size $75 \times 75$. The input consists of a question and an image. The task contains 3 types of questions - Unary, Binary, and Ternary. Unary questions consider properties of single objects. For example, \textit{what is the color of the square?}. Binary questions consider relations between two objects. For example, \textit{what is the shape of the object closest to the red object?}. Similarly, ternary questions consider relations between 3 obejcts.  
The Sort-of-CLEVR images consist of 6 randomly placed geometrical shapes of 6 possible colors and 2 possible shapes. 

To demonstrate the generality of CNT, we evaluate both convolutional resnets (Preactresnet18 with 64 initial channels) and transformers on the Sort-of-CLEVR benchmark.  Unlike with equilateral shape classification and image classification, we used relu everywhere in the convolutional network for the Sort-of-CLEVR tasks.  We also injected a question-embedding along with the label and noise-level embeddings as a way of conditioning on the question.  We report results on Sort-of-CLEVR (Table~\ref{tab:sortofclevr}).  Additionally, we found that accuracy converged fastest at the lowest noise levels (Figure~\ref{fig:shapes_noise}) and more gradually at higher noise levels.

\begin{table}
    \small
    \vspace{-2mm}
	\caption{Sort-Of-Clever (Ternary: question relating to three object, Binary: question relating to two object, Unary: question relating to one object): Test accuracy mean (standard deviation) from 5 seeds}
	\label{tab:sortofclevr}
	\centering
	\resizebox{.45\textwidth}{!}{%
	\begin{tabular}{cccc}
		\toprule
		Model & Ternary & Binary & Unary  \\
		\cmidrule(){1-4}
		\multicolumn{4}{c}{ResNet - ReLU \& Batch-norm} \\
		\cmidrule(){1-4}
	     Baseline & 74.00 (1.49) & 78.34 (8.38) & 99.93 (0.09) \\
	     only-noise & {\fontseries{b}\selectfont 76.86} (1.52) & 85.17 (6.56) & {\fontseries{b}\selectfont 100.0} (0.00) \\
	     CNT & 74.80 (2.38) & {\fontseries{b}\selectfont 90.73} (1.30) & {\fontseries{b}\selectfont  100.0} (0.00) \\
		\cmidrule(){1-4}
		\multicolumn{4}{c}{ResNet - Mish \& Batch-norm} \\
		\cmidrule(){1-4}
	     Baseline & 75.60 (3.52) & {\fontseries{b}\selectfont 91.93} (2.39) & 93.17 (6.63) \\
	     only-noise & {\fontseries{b}\selectfont 76.27} (2.98) & 89.08 (5.67) & 97.90 (2.37) \\
	     CNT & 73.80 (3.03) &  86.49 (8.35) & {\fontseries{b}\selectfont 99.63} (0.66) \\
		\cmidrule(){1-4}
		\multicolumn{4}{c}{ResNet - ReLU \& Group-norm} \\
		\cmidrule(){1-4}
	     Baseline & {\fontseries{b}\selectfont 83.72} (0.70) & 97.43 (0.11) & {\fontseries{b}\selectfont 100.0} (0.00) \\
	     only-noise & 83.64 (0.59) & 97.77 (0.16) & 99.98 (0.05) \\
	     CNT & 82.44 (0.86) & {\fontseries{b}\selectfont 97.91} (0.23) & {\fontseries{b}\selectfont  100.0} (0.00) \\
		\cmidrule(){1-4}
		\multicolumn{4}{c}{ResNet - Mish \& Group-norm} \\
		\cmidrule(){1-4}
	     Baseline & 83.15 (0.81) & 97.84 (0.20) & {\fontseries{b}\selectfont  100.0} (0.00) \\
	     only-noise & {\fontseries{b}\selectfont 83.72} (0.59) & {\fontseries{b}\selectfont 97.89} (0.29) & {\fontseries{b}\selectfont 100.0} (0.00) \\
	     CNT & 82.81 (0.80) & 97.49 (0.68) & {\fontseries{b}\selectfont  100.0} (0.00) \\
		\cmidrule(){1-4}
		\multicolumn{4}{c}{Transformers} \\
		\cmidrule(){1-4}
	     Baseline & {\fontseries{b}\selectfont 54.40} (0.80) & 72.20 (2.48) & 78.40 (11.60) \\
	     only-noise & 55.6 (2.87) & {\fontseries{b}\selectfont 77.0} (2.45) & {\fontseries{b}\selectfont 98.2} (0.40) \\
	     CNT & 52.40 (0.49) & 73.00 (5.02) & 83.60 (12.60) \\
		\bottomrule
	\end{tabular}
	}
	\vspace{-3mm}
\end{table}




\subsection{Reinforcement Learning with Decision Transformers}

We test CNT on the atari benchmark \cite{atari} following the same setup as decision transformer \cite{DBLP:journals/corr/abs-2106-01345}. The key in decision transformer is the choice if trajectory representation. The trajectory is represented such that it allows conditional generation of actions based on future expected rewards. This is done by conditioning the model on return-to-go $\hat{R}_k = \sum_{k'=k}^K r_k$, where $k$ denotes the timesteps. This results in the following trajectory representation: $\tau = \big( \hat{R}_1, s_1, a_1, \hat{R}_2, s_2, a_2, \hat{R}_3, s_3, a_3, \ldots \big)$.  At test time, we can condition the desired return $\hat{R}_1$ and the start state $s_1$ to generate actions. 


Similar to decision transformer, we use a fixed context of length of $K$ to train our models i.e. we only feed in the last $K$ timesteps. This results in a sequence length of $3K$ (considering 3 modalities - returns, states, and actions). Each modality is processed into an embedding - The states are processed using a convolutional encoder into an embedding, the returns and actions are processed using linear layers. The processed tokens are fed into a GPT \cite{Radford2018ImprovingLU} model. The outputs corresponding to $s_k$ are fed into a linear layer to predict the action $a_k$ to be taken at $k$. To incorporate CNT, we replace the layer normalization layers in GPT with conditional normalization layers. During training, we feed the noisy actions $a_k(t)$ and $t_k$ through the conditional normalization layers and the model is tasked with predicting $a_k(0)$ (the true action). 

We train on 500000 examples from DQN replay dataset similar to \citep{DBLP:journals/corr/abs-1907-04543}. We run on 4 atari games - Qbert, Seaquest, Pong, and Breakout. We use a context length ($K$) of 30. We report the mean and variance across 10 seeds. We report the results in Table \ref{tab:atari_results}.

\begin{table}
	
\footnotesize	

\centering
    \caption{Here we report the returns obtained by the models on the four Atari Games studied in the Decision Transformer paper \citep{DBLP:journals/corr/abs-2106-01345}. Mean and standard deviation are reported across 10 seeds.}
    \scriptsize	
\centering 
	\resizebox{.45\textwidth}{!}{%
    \begin{tabular}{ccc}
    \hline
        Model & Baseline & CNT \\
         \hline
        Qbert & 3084.22 (1560.93) & \textbf{3390.22 (2360.71)} \\
        Seaquest & 894.67 (368.57) & \textbf{1106.75 (148.74)} \\
        Breakout & \textbf{74.67 (14.55)} & 63.22 (9.60) \\
        Pong & \textbf{11.22 (6.16)} & 9.22 (6.78) \\
        \hline
    \end{tabular}}
    
    \label{tab:atari_results}
\end{table}

\subsection{Injecting Laplace Noise}

\begin{figure}[h!]
    \centering
    \includegraphics[width=1\linewidth]{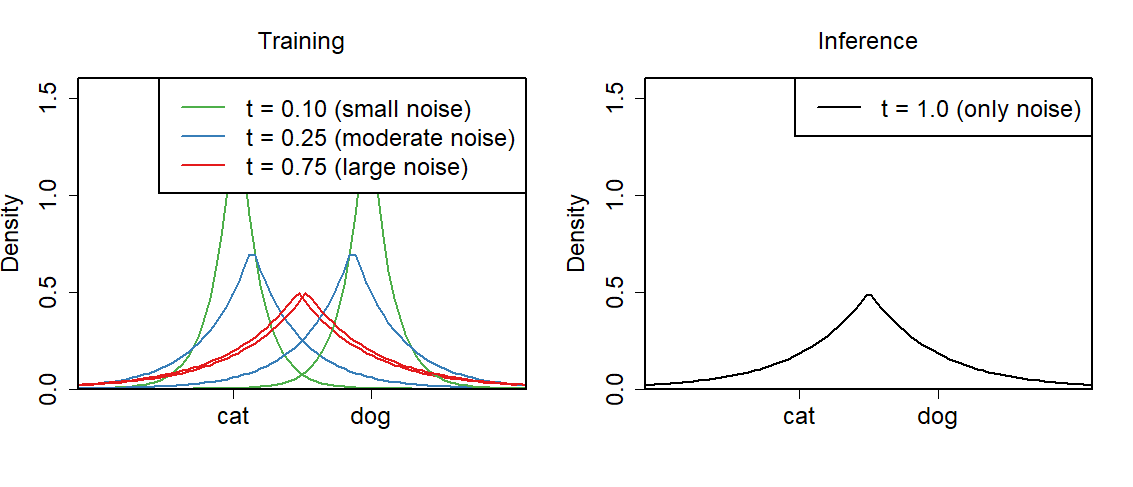}
    \caption{This figure illustrates CNT in a ’cat’ versus ’dog’ classification problem with the Laplace distribution instead of the Gaussian distribution.
    When the noise is small, the model can guess that the true label is 'cat'. When the noise is moderate, if the label is left from 'cat' it remain clear that the true label is 'cat', otherwise, it becomes more difficult to tell and the model must more focus on the input image. When the noise is large, the true label could be anything, thus the model will ignore the noisy label and focus entirely on the input image.
    } 
  \label{fig:laplace_noise} 
  \vspace{-4mm}
\end{figure}

To demonstrate the generality of CNT, we also trained with noisy targets where the noise follows a Laplace distribution instead of a Gaussian distribution (Figure~\ref{fig:laplace_noise}).  Aside from this change, we kept all hyperparameters the same and used the Mish activation.  These results are shown in Table~\ref{tab:laplace}.  

\begin{table}
\footnotesize	
\centering
    \caption{We report classification accuracy to compare CNT with gaussian noise and CNT with Laplace noise.  All methods use the Mish activation and PreActResnet18 architecture).}
    \scriptsize	
\centering 
	\resizebox{.45\textwidth}{!}{%
    \begin{tabular}{cccc}
    \hline
        Model & Baseline & Gaussian & Laplace \\
         \hline
        CIFAR-10 & 96.74 & 96.70 & \textbf{96.94} \\
        CIFAR-100 & 79.21 & \textbf{80.23} & 78.54 \\
        Tiny-Imagenet & 63.77 & 65.32 & \textbf{65.68} \\
        \hline
    \end{tabular}}
    
    \label{tab:laplace}
\end{table}

\section{Limitations}

We tested our approach on a wide variety of tasks to verify its general validity. However, we only tested one way of conditioning on the noisy target, which is not necessarily the best.

Progressive learning has known theoretical benefits on generalization. However, there is less theoretical evidence (as far as we can tell) for the benefits of multi-scale methods on generalization. It is possible that using a specific annealing process to decrease the noise during training in a progressive learning fashion would also provide good improvements in generalization. However, we did not test this idea as we prefer the multi-scale approach's simplicity, which does not require any hyper-parameter tuning. Our goal is to have a simple regularization method that does not require tuning.

We injected noise directly into the target space (for example, one-hots for classification problems), which does not reflect the semantic structure of the problem being solved.  We used this approach because it is fully general and does not assume any special knowledge.  However, it would also be interesting to inject noise into a semantically meaningful space instead of the original target space.  

We used Gaussian noise due to its simplicity and the fact that the VP process has obtained great success in score-based generative models. We also used Laplace noise. However, other types of noise could be tried. 

\section{Conclusion}

We devised a new regularizer called CNT, which leverages top-down feedback and multi-scale learning by conditioning the neural network on a noisy target. This noisy target acts as a hint that allows the model to succeed faster on easier tasks with less noise than harder ones. CNT thus provides a continuous path for the model to progress from easy to hard examples. We show that we obtain higher generalization when using CNT on a wide variety of different tasks. The improvement from using CNT is particularly significant in reinforcement learning and when the model has low capacity. This suggests that conditioning on a noisy target is especially beneficial for difficult problems. 

Notably, only conditioning on pure noise also offered a slight benefit; this suggests that conditioning on noise may be beneficial on its own. We hypothesize that the network still attempts to solve the problem as multiple sub-problems (conditional on the value of $t$) even though the problem is the same at every value of $t$; this may regularize the network and reduce overfitting. However, given the more substantial theoretical and empirical evidence for CNT, it is more sensible to condition on the noisy target than simply on noise when conditioning on the noisy target is no more difficult.




\clearpage
\bibliographystyle{unsrtnat}
\interlinepenalty=10000
\bibliography{paper}

\end{document}